\pdfoutput=1

\documentclass[11pt]{article}

\usepackage{acl}

\usepackage{times}
\usepackage{latexsym}

\usepackage{longtable}
\usepackage{graphicx}
\usepackage{subfigure}
\usepackage{bm}
\usepackage{amsmath}
\usepackage{bbm}
\usepackage{multirow}
\usepackage{array, caption, tabularx, makecell}
\setcellgapes{3pt}
\usepackage{url}            
\usepackage{booktabs}       
\usepackage{amsfonts}       
\usepackage{nicefrac}       
\usepackage{microtype}      
\usepackage{xcolor}   

\usepackage[T1]{fontenc}

\usepackage[utf8]{inputenc}

\usepackage{microtype}

\title{Diversifying Neural Text Generation with Part-of-Speech Guided Softmax and Sampling}

\author{Zhixian Yang \and Pengxuan Xu \and Xiaojun Wan \\
  Wangxuan Institute of Computer Technology, Peking University \\
  Center for Data Science, Peking University \\
  The MOE Key Laboratory of Computational Linguistics, Peking University \\
  \texttt{yangzhixian@stu.pku.edu.cn} \\
  \texttt{xupengxuan@stu.pku.edu.cn} \\
  \texttt{wanxiaojun@pku.edu.cn} \\}

\begin{document}
\maketitle
\begin{abstract}
Neural text generation models are likely to suffer from the low-diversity problem. Various decoding strategies and training-based methods have been proposed to promote diversity only by exploiting contextual features, but rarely do they consider incorporating syntactic structure clues. In this work, we propose using linguistic annotation, i.e., part-of-speech (POS), to guide the text generation. In detail, we introduce POS Guided Softmax to explicitly model two posterior probabilities: (i) next-POS, and (ii) next-token from the vocabulary of the target POS. A POS Guided Sampling strategy is further proposed to address the low-diversity problem by enriching the diversity of POS. Extensive experiments and human evaluations show that, compared with existing state-of-the-art methods, our POS Guided Softmax and Sampling (POSG) can generate more diverse text while maintaining comparable quality.\footnote{Our code is available at \url{https://github.com/FadedCosine/POS-Guided-Neural-Text-Generation}}
\end{abstract}

\section{Introduction}

Maximum likelihood estimation (MLE) is a standard approach to training a neural text generation model, e.g. Transformer \cite{attention}, to generate human-like text. However, existing generation systems often suffer from the low-diversity problem \cite{topp, unlikelihood}, which leads to dull and repetitive generations. This problem unavoidably affects the overall generation quality. 

We conclude that the low-diversity problem is mainly manifested in two aspects: \textit{form} and \textit{content} \cite{Fu_Lam_So_Shi_2021, topp, diversity-eval}. As shown Table~\ref{diversity-example}, the low form diversity can be reflected in repeating some words, using similar lexicon and syntax, and more. The low content diversity can be expressed as a single and dull content with nothing different.

\begin{table*}[t]
	\small
	\begin{center}
	\resizebox{\textwidth}{!}{
		\begin{tabular}{p{\textwidth}}
			\toprule
			\textbf{Context}: The NK 2nd Division , concentrated in the Sinban-ni area west of the river , had , in effect , attacked straight east across the river and was trying to seize the two avenues of advance into Changnyong. \\
			\hline
			\hline
			\textbf{Text 1}: They were joined by \textcolor{blue}{the 27th Battalion, US 24th Infantry Regiment, the 27th Regiment, and the 27th Regiment.} \textcolor{blue}{The US 24th Infantry Division}, under the command of Major General John R. Dempsey, was ordered to charge \textcolor{blue}{the US 24th Infantry Division}, and capture it from the west.  \\
			\hline
			\textbf{Text 2}: 
		\textcolor{orange}{The NK} 2nd \textcolor{orange}{Division}, \textcolor{orange}{which had been} involved in the assault, was forced to withdraw from the area north of the river.
		 \textcolor{orange}{The NK} 3rd \textcolor{orange}{Division, which had been fighting in the area since the} beginning \textcolor{orange}{of the battle,} was moved to the south. \textcolor{orange}{The NK} 2nd \textcolor{orange}{Division, which had been fighting in the area since the} start \textcolor{orange}{of the battle,} had been pushed back to the south.\\
			\hline
			\textbf{Text 3}: The 2nd Division had been moving north from Alcester's position on the road, but were not expecting another attack. In the immediate aftermath of the attack, to keep the 2nd Division in reserve, which had been preparing for an attack on Hill 131. Along with the 3rd Battalion of the US 2nd Infantry Regiment, attacked Hill 129 at Pakchon on the way to Beaulieu.\\
			\bottomrule
		\end{tabular}
	}
	\end{center}
		\caption{Examples of low-diversity generated text, given context from the Wikitext-$103$ dataset \cite{WIKITEXT-103}. Text 1 has a poor form diversity due to many useless repeating words (highlighted in \textcolor{blue}{blue}). Text 2 keeps talking about only one single content, with similar lexicon and syntax (highlighted in \textcolor{orange}{orange}), indicating low diversity in both terms of form and content. Though Text 3 has various syntactical and lexical forms with no repetition, all the content of it is about the ``attacks'', which means low content diversity. Text 1 is sampled from MLE, Text 2 from F$^2$-Softmax \cite{F2-Softmax}, and Text 3 from FACE \cite{face} (Section~\ref{Setup}).}
	\label{diversity-example}
\end{table*}

Several feasible fixes have been proposed, such as post-hoc sampling strategies including temperature \cite{temperature}, top-$k$ \cite{topk}, and nucleus sampling \cite{topp}. Recently, some works suggest that it is the maximizing likelihood itself that should account for the low-diversity problem \cite{topp, unlikelihood}. \citet{topp} think that MLE can not adequately capture the rich diversity and expression in human language. \citet{F2-Softmax} argue that the imbalanced token distribution inherent in natural language even worsens the low-diversity problem. Based on these analysis, many training-based methods have been proposed. \citet{unlikelihood} propose the unlikelihood training to penalize repetition with auxiliary losses. \citet{face} propose to utilize dynamically scaling losses conditioned on the token frequency in the training phase. \citet{F2-Softmax} factorize the probability distribution and design an elaborate token cluster algorithm for a balanced training. 

Though those encouraging progress has been made, we argue that current training-based methods only take plain contextual features to promote diversity, rarely considering incorporating syntactic structure clues. For example, when humans are writing articles, it is natural to predetermine the part-of-speech (POS) before giving the next token. Existing studies have verified that incorporating POS can improve the translation quality in neural machine translation (NMT) \cite{POSNMT, POS-NAT}. Intuitively, since the vocabularies of different POS vary a lot, the diversity of POS will certainly lead to the diversity of text. Unfortunately, we observe that existing methods with no consideration of the inner POS structure fail to learn the diversity of POS in human language (shown in Table~\ref{tab-LM-ppmcc}).

All these factors motivate us to address the low-diversity problem with the guidance of POS. Thus, in this work, we first present the POS Guided Softmax (Figure~\ref{fig-POSSoftmax}), building upon a hybrid decoder that predicts two posterior probabilities: (i) next-POS, and (ii) next-token from the vocabulary of the target POS. Our work shows that, following the POS clue, our model can gain a deeper insight into text's syntactic structure. Thereafter, we propose a POS Guided Sampling to improve the diversity of generated text lexically and syntactically while maintaining comparable quality.

To sum up, the contributions of our work are three-fold. (i) We introduce a novel POS Guided Softmax, incorporating POS tags as the observed discrete decisions to improve text generation. (ii) Based on POS Guided Softmax, POS Guided Sampling is proposed to promote text diversity effectively without degrading quality. (iii) We conduct extensive experiments on language modeling and paraphrase generation. Experimental results and human evaluation show that our model can easily adapt to different downstream tasks and generate text with high diversity as well as quality.  

\section{Related Works}
\subsection{Diversity-promoting Methods}

\noindent\textbf{Decoding-based Methods.} Although greedy search and beam search are well known decoding strategies for neural text generation, \citet{topp} have shown that these methods always generate generic, repetitive, and awkward words. \citet{Kulikov2019ImportanceOS} and \citet{diverse-beam} have proposed several variants of beam search as alternatives. Recently, stochastic decoding methods have been widely used, and some studies propose to sample from a truncated and renormalized Softmax distribution. Top-$k$ sampling \cite{topk} only samples from the top-$k$ most probable tokens. Nucleus sampling \cite{topp} only samples from the smallest set whose cumulative probability is at least $\alpha$. However, those decoding-based methods are lack of controllability. Combined with above methods, our proposed method can further promote diversity using POS as a more controllable clue.

\noindent\textbf{Training-based Methods.}
As a standard approach to training a neural text generation model, MLE has been proved to be defective. \citet{F2-Softmax} have shown that MLE may mislead the model because of the imbalanced token distribution. Thus, they design a greedy approach MefMax and factorize Softmax to ensure a balanced training according to the word frequency. FACE \cite{face} utilizes the target word frequency to modify the cross-entropy loss with a frequency-based weight factor. \citet{unlikelihood} introduce an unlikelihood loss to implicitly reduce the frequent tokens and potential repeats. Other approaches, such as negative training \cite{NegativeTraining}, reinforcement learning \cite{ReinforcementDiversity}, and imitation learning \cite{ImitationDiversity}, have recently been applied to promote the diversity during the training phase. All above training-based methods only learn from plain contextual features, while ignoring other linguistic features. Our focus is on leveraging POS features to guide both phases of training and decoding.

\subsection{POS in Text Generation}
Previous works, which leverage POS for text generation, can be summarized as follows:


\textbf{POS in Encoding}. A branch of previous works \cite{ICPOSEncoding, WMTSennrichH16, WrayCLD19} explore to adopt POS on the encoding side to help language understanding and generation. \citet{WMTSennrichH16} concatenate the embeddings of POS tags with sentence features to improve the translation quality. For the image caption generation, \citet{ICPOSEncoding} use POS tags to control the fusion of the image features and the related word embeddings.  \citet{WrayCLD19} enrich the encoding with POS of the accompanying captions for cross-modal search tasks.

\textbf{POS in Decoding}. The second line of studies directly model the POS structure during decoding. \citet{SuLYC18} introduce a hierarchical decoder that relies on teacher forcing to learn different POS patterns on different layers. \citet{DeshpandeAWSF19} use POS tag sequences as summaries to implicitly drive image caption generation. \citet{YangLXWB19} treat POS tags as latent variables in NMT and optimize the model by Expectation Maximization (EM). \citet{POS-NAT} employ POS sequences to constrain the non-autoregressive generation (NAG) modes to alleviate the multi-modality problem. However, all the previous studies only focus on a single specific task and leverage POS as hidden decoding features \cite{DeshpandeAWSF19, YangLXWB19}, teacher forcing techniques \cite{SuLYC18, BugliarelloE21} or NAG plannings \cite{POS-NAT} in order to improve the generic quality of generated texts, while our proposed methods regard POS tags as observed sequential variables and directly model the POS distribution during both phases of training and decoding with the goal of improving text diversity. 

To our best knowledge, we are the first to introduce an explicit POS-guided generation method as a generic way to promote text diversity while maintaining quality.


\section{Language Modeling}

The goal of language models is to assign a probability to text (i.e. word sequence) $\mathbf{x} = \left[ x_1, \dots, x_T\right] $, where each $x_t$ in the sequence is a token from a vocabulary $\mathcal{V}$, i.e., $x_t \in \mathcal{V}$, and $T \in \mathbb{N}$. We train the language models to learn a distribution $p_\theta\left(\mathbf{x}\right)$ with the goal to fit the ground-truth distribution $p_\star\left(\mathbf{x}\right)$ for all $\mathbf{x}$. Specifically, when the language model is a neural network, $\theta$ is regarded as the model parameters of the neural network, and we can factorize $p_\theta\left(\mathbf{x}\right)$ as $p_\theta\left(\mathbf{x}\right) = \Pi_{t=1}^{T} p_\theta\left(x_t \mid \mathbf{x}_{<t}\right) $. The conventional approach for learning the language model parameters $\theta$ is to maximize the log-likelihood by minimizing: 
\begin{equation}
\begin{aligned}
	\label{mle}
	\mathcal{L}_{\mathrm{MLE}}\left(\theta\right) =&-\sum_{t=1}^{T} \log p_{\theta}\left(x_{t} \mid \mathbf{x}_{<t}\right), \\
	p_{\theta}\left(x_t \mid \mathbf{x}_{<t}\right)&=\frac{\exp \mathbf{h}_{t-1}^{\top} \mathbf{w}_{x_t}}{\sum_{x \in \mathcal{V}} \exp \mathbf{h}_{t-1}^{\top} \mathbf{w}_{x}},
\end{aligned}
\end{equation}
where $\mathbf{h}_{t-1}$ is a hidden state of the context $\mathbf{x}_{<t}$, and $\mathbf{w}_{x_t}$ is the output embedding vector for $x_t \in \mathcal{V}$. 


\section{Methodology}
In this section, we describe an overview of our
proposed method, POS Guided Softmax and Sampling (POSG). POSG is designed to exploit syntactic structure, i.e., POS tags for text generation in both the training and decoding phases. Specifically, giving text sequence $\mathbf{x} = [ x_1, \dots, x_T] $, we first use off-the-shelf POS tagger \cite{StanfordCoreNLP} to annotate corresponding POS sequence $\bm{\rho} = [ \rho_1, \dots, \rho_T] $, where each $\rho_t$ is a POS tag from the POS vocabulary $\mathcal{P}$, i.e., $\rho_t \in \mathcal{P}$, and $T \in \mathbb{N}$. We define all the tokens whose POS is $\rho$ as a vocabulary $\mathcal{V}_{\rho}$, where $\mathcal{V}_{\rho} \subset \mathcal{V}$.

\subsection{POS Guided Softmax}
\begin{figure*}[t]
	\begin{center}
		\centerline{\includegraphics[width=0.75\textwidth]{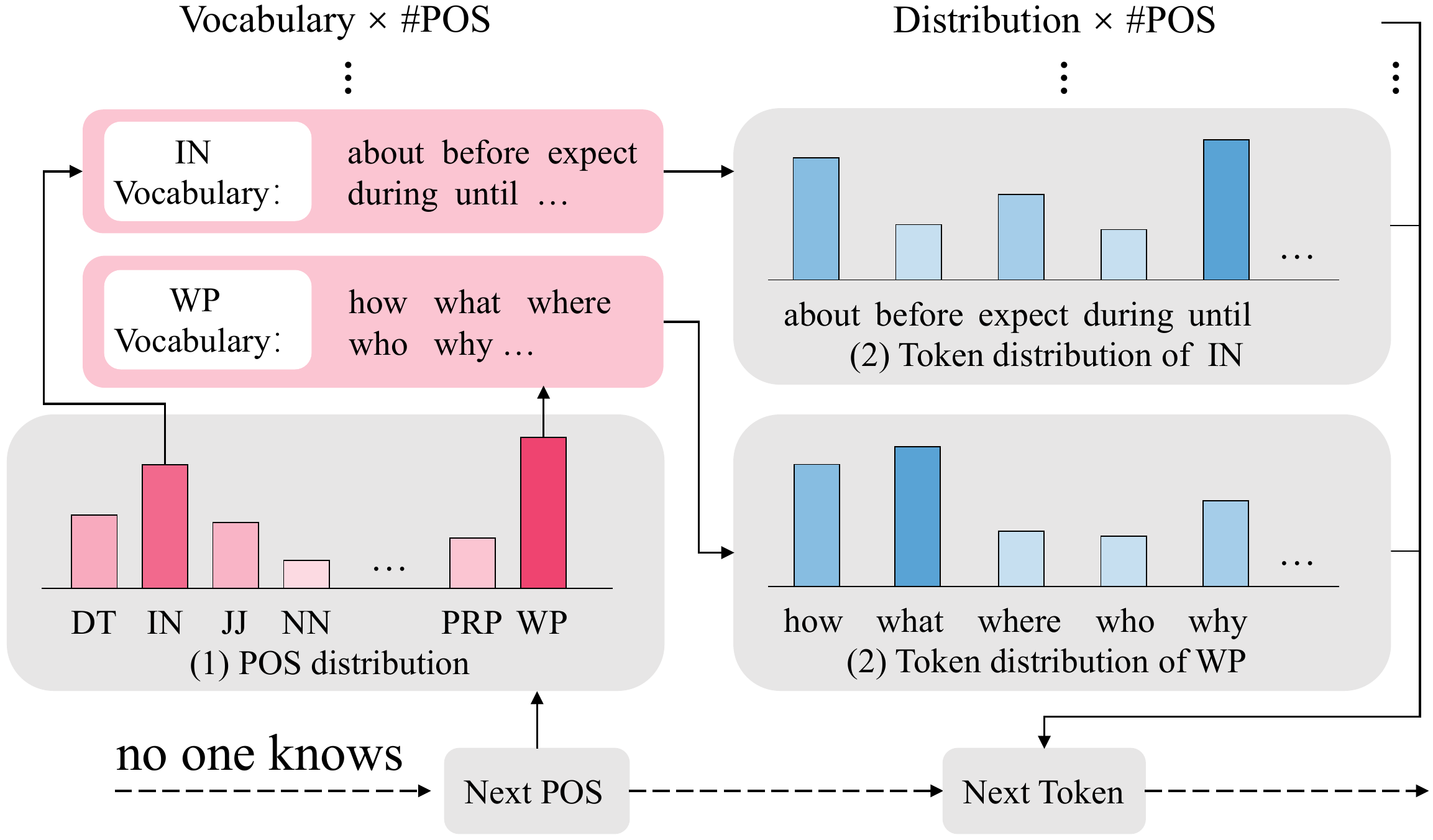}}
		\caption{Illustration of POS Guided Softmax. The decoding process is decomposed into two stages: first predicts the next-POS distribution, and then decodes the next-token distribution from the vocabulary of the previously predicted POS. Since there exist some tokens with more than one POS, the final next-token distribution is the sum of all the POS's token distributions.}
		\label{fig-POSSoftmax}
	\end{center}
	\vskip -0.25in
\end{figure*}

Figure~\ref{fig-POSSoftmax} illustrates the core idea of our POS Guided Softmax. Given a context, there exist various choices for the next POS, which can be modeled as the next POS distribution. For the context \textit{``no one knows''}, the next possible POS includes WH-pronoun (WP), preposition (IN), etc. For example, if WP is predicted as the next POS, the model will decode the next token from the WP vocabulary ($\mathcal{V}_{\text{WP}}$) with the token distribution of WP. Consequently, the complete sequence can be \textit{``no one knows what will happen''}. For another case, if IN is predicted as the next POS, the next token will be decoded from $\mathcal{V}_{\text{IN}}$ with the corresponding token distribution. Then, the sequence may end up saying \textit{``no one knows until it finally happens''}. This example also shows that the different choices of POS at each time step can result in vastly different generated text, thus promoting text diversity.

Following the core idea, we assume that the decoding process can be divided into two stages: for each time $t$, a POS tag $\rho_t$ is predicted first, and then the model decodes next-token $x_t$ from $\mathcal{V}_{\rho_t}$. Therefore, the joint conditional probability of $x_t$ and its corresponding POS tag  $\rho_t$ is formulated as:
\begin{equation}
	\begin{aligned}
		\label{joint-p}
		p_{\theta} \left(x_{t}, \rho_t \mid \mathbf{x}_{<t}\right) &=p_{\theta_1}\left(\rho_{t} \mid \mathbf{x}_{<t} \right) \\
		& \times p_{\theta_2}\left(x_{t} \mid \rho_{t}, \mathbf{x}_{<t}\right),
	\end{aligned}
\end{equation}
where $p_{\theta_1}\left(\rho_{t} \mid \mathbf{x}_{<t}\right)$ is the next-POS probability and $ p_{\theta_2}\left(x_{t} \mid \rho_{t}, \mathbf{x}_{<t}\right)$ is the next-token probability conditioned on $\rho_{t}$. These probabilities are defined empirically by applying a linear output embedding on $\mathbf{h}_{t-1}$ and then a Softmax function respectively:
\begin{equation}
\small
\setlength{\abovedisplayskip}{5pt}
\setlength{\belowdisplayskip}{3pt}
	\begin{aligned}
		\label{POSSoftmax}
		p_{\theta_1}\left(\rho_{t} \mid \mathbf{x}_{<t}\right) &=\frac{\exp \mathbf{h}_{t-1}^{\top}\mathbf{o}_{\rho_{t}}}{\Sigma_{\rho \in \mathcal{P}} \exp \mathbf{h}_{t-1}^{\top}\mathbf{o}_{\rho}},  \\
		p_{\theta_2}\left(x_{t} \mid \rho_{t}, \mathbf{x}_{<t}\right) &=
		\begin{cases}
			\frac{\exp \mathbf{h}_{t-1}^{\top}  \mathbf{w}_{x_{t}}} {\Sigma_{x \in \mathcal{V}_{\rho_{t}}} \exp \mathbf{h}_{t-1}^{\top}  \mathbf{w}_{x}}, & \text{if } x_t \in \mathcal{V}_{\rho_{t}} \\
			0, & \text{otherwise }
		\end{cases},
	\end{aligned}
\end{equation}
where $\mathbf{o}_{\rho_t}$ and $\mathbf{w}_{x_t}$ are the output embeddings for $\rho_t \in \mathcal{P}$ and $x_t \in \mathcal{V}_{\rho_t}$, respectively. In this way, we regard POS tags as observed sequential variables, which also contributes to the model interpretability and controllability. Then, the final next-token distribution can be formulated as: $
	p_{\theta}\left(x_t \mid \mathbf{x}_{<t}\right)=\sum_{\rho_t \in \mathcal{P}} p_{\theta} \left(x_{t}, \rho_t \mid \mathbf{x}_{<t}\right).$
Note that some tokens may have more than one POS, and $p_{\theta} \left(x_{t}, \rho_t \mid \mathbf{x}_{<t}\right) = 0$ for $x_{t} \notin \mathcal{V}_{\rho_{t}} $. Since the number of POS in a specific language family is fixed, there is no problem of insufficient exploration in variables' space.

As mentioned before, we think of POS tags as observed sequential variables and extend the training text set with annotated POS sequences, so we define the POS guided training objective as follows:
\begin{equation}
\setlength{\abovedisplayskip}{-5pt}
\setlength{\belowdisplayskip}{3pt}
	\begin{aligned}
		\label{POSobj}
		\mathcal{L}_{\text{POS-Guided}}\left(\theta\right) 
		=&-\sum_{t=1}^{T} \big[\log p_{\theta_1}\left(\rho_{t} \mid \mathbf{x}_{<t}\right) \\
		&+ \log p_{\theta_2}\left(x_{t} \mid \rho_{t}, \mathbf{x}_{<t}\right) \big].
	\end{aligned}
\end{equation}

\subsection{POS Guided Sampling}
We propose POS Guided Sampling based on POS Guided Softmax. Consistent with POS Guided Softmax, the key idea is to divide the whole sampling process into two stages: \textit{POS sampling} and \textit{token sampling}. In POS sampling, we first sample a POS, and then in token sampling, we use the sampled POS to control the sampling of tokens. Note that arbitrary sampling strategies can be adopted to both the POS sampling and token sampling. Here, we take top-$k$ sampling for POS sampling, and nucleus sampling for token sampling as an example, and then we can formulate our POS Guided Sampling as follows:
\begin{equation}
\small
    \begin{aligned}
		\label{eq-posg-sampling}
		p^{\prime}_{\theta}\left(x_t \mid \mathbf{x}_{<t}\right) =\sum_{\rho_t \in \mathcal{P}} & \big[ p^{\prime}_{\theta_1} \left( \rho_t \mid\mathbf{x}_{<t}\right)  \times  p^{\prime}_{\theta_2} \left(x_{t} \mid \rho_t , \mathbf{x}_{<t}\right) \big], \\
		p^{\prime}_{\theta_1} \left(\rho_t \mid \mathbf{x}_{<t}\right) &=
		\begin{cases}
			\frac{p_{\theta_1} \left(\rho_t \mid \mathbf{x}_{<t}\right) }{ \mathcal{Z}_{\theta_1}},&  \text{ if } \rho_t \in \mathcal{P}^{\prime} \\
			0,&  \text{ otherwise }
		\end{cases}, \\
		p^{\prime}_{\theta_2} \left(x_t \mid \rho_t, \mathbf{x}_{<t}\right) &= 
		\begin{cases}
			\frac{p_{\theta_2} \left(x_t \mid \rho_t , \mathbf{x}_{<t}\right) }{ \mathcal{Z}_{\theta_2}}, & \text{ if } x_t \in \mathcal{V}^{\prime}_{\rho_{t}}\\
			0,&   \text{otherwise }
		\end{cases},\\
		\mathcal{Z}_{\theta_1} &= \sum_{\rho \in \mathcal{P}^{\prime}} p_{\theta_1} \left(\rho \mid \mathbf{x}_{<t}\right), \\
		\mathcal{Z}_{\theta_2} &= \sum_{x \in \mathcal{V}^{\prime}_{\rho_{t}}}p_{\theta_2} \left(x \mid \rho_t , \mathbf{x}_{<t}\right), 
    \end{aligned}
\end{equation}
where $\mathcal{P}^{\prime} \subset \mathcal{P} $ is a POS set containing top-$k$ most probable POS tags, and $\mathcal{V}^{\prime}_{\rho_{t}} \subset \mathcal{V}_{\rho_{t}} $ is the smallest token set such that $
\sum_{x \in \mathcal{V}^{\prime}_{\rho_{t}} }  p_{\theta_2}\left(x \mid \rho_{t} , \mathbf{x}_{<t}\right) \geq \alpha^{\text{(token)}}$. $k^{\text{(POS)}}$ and $\alpha^{\text{(token)}}$ ($0 < \alpha^{\text{(token)}} \leq 1$) are the hyper-parameters for the sampling of POS and token, respectively. For other sampling strategies used in POS sampling and token sampling, POS Guided Sampling can be similarly defined. 

\section{Experiments}

We systematically evaluate our proposed methods on language modeling task (Section~\ref{exper-lm}) and paraphrase generation task (Section~\ref{exper-para}).
\subsection{Experimental Setup}
\label{Setup}
\noindent\textbf{Model Architecture}\quad Since our proposed methods are architecture agnostic, we implement POS Guided Softmax on the Transformer \cite{attention}, a widely used architecture for neural text generation.
Details of the experimental setup for each task are shown in Appendix~\ref{DetailSetup}.

\noindent\textbf{Baseline Models}\quad We compare our \textbf{POS Guided Softmax and Sampling (POSG)} with the following baselines: (i) \textbf{Maximum likelihood estimation (MLE)}, a standard approach for neural text generation. (ii) \textbf{Frequency-Aware Cross-Entropy (FACE)} \cite{face} dynamically weights the cross-entropy losses conditioned on the token frequency. (iii) \textbf{Frequency Factorization Softmax (F$^2$-Softmax)} \cite{face} factorizes the standard Softmax based on the token frequency. (iv) \textbf{Unlikelihood training (UL)} \cite{unlikelihood} is to enhance the log-likelihood loss with an unlikelihood loss that penalizes the generation of repeated tokens. (v) We further implement two task-specific baselines: \textbf{Mixture of Softmaxes (MoS)} \cite{MoS} for language modeling,  \textbf{Syntax Guided Controlled Paraphraser (SGCP)} \cite{SGCP} for paraphrase generation. Note that decoding-based methods, including top-$k$ and nucleus sampling, can be directly compared to POSG, when they are applied to MLE. The details will be described in the sections of Generation Details.  

\subsection{Language Modeling}
\label{exper-lm}

\begin{table*}[t]
	\centering
	\resizebox{\textwidth}{!}{
		\begin{tabular}{l|cccccc|ccccc}
\toprule
\multirow{2}{*}{Models}    & \multirow{2}{*}{Self-BLEU4 $\downarrow$} & \multirow{2}{*}{Rep $\downarrow$} & \multirow{2}{*}{Uniq $\uparrow$} & \multicolumn{3}{c|}{Distinct $\uparrow$}     &  \multirow{2}{*}{PPL $\downarrow$} & \multirow{2}{*}{KLD $\downarrow$} & \multicolumn{3}{c}{MS-Jaccard $\uparrow$}                 \\
  & & & & n=1   & n=2    & n=3   &     &  & n=1 & n=2  & n=3       \\ \hline
MLE         & 46.9                       & 1.86                 & 11.7k                                      & 50.2          & 77.2          & 86.2       & \textbf{32.7}        & 1.34                 & 56.9          & 38.2          & 25.4  \\
FACE   & \underline{34.2}                                             & 1.56                 & \textbf{14.9k}                                      & 60.0          & 85.1          & 90.6           & 36.1   & \underline{1.18}   & 58.6   & 37.6  & 24.0            \\
F$^2$-Softmax           & 51.5     & 4.09                 & 10.8k                                      & 42.4          & 65.3          & 75.2     & 35.0  & 1.58 & 51.5  & 33.7          & 22.4    \\
UL    & 42.4                                             & \underline{0.240}                 & 12.8k                                      & \textbf{61.2} & \underline{87.8}   & \underline{93.3}   & 37.0  & 1.20  & \underline{61.2}          & \underline{40.4}          & \textbf{26.2} \\ 
MoS  & 55.3 & 3.99 & 8.40k & 48.2 & 74.3 & 83.0   & 38.2 & 1.48 & 56.9 & 38.1 & 25.4 \\
\hline
POSG       & \textbf{34.1}       &  \textbf{0.000}                 & \underline{13.8k}            & \underline{60.2}  & \textbf{88.8} & \textbf{94.3}     &  \underline{34.4}   & \textbf{1.17}        & \textbf{62.2} & \textbf{40.7} & \underline{25.9}     \\
\bottomrule
		\end{tabular}
	}
\caption{Automatic evaluation results for different models on the language modeling task. Numbers $n \in \{1, 2, 3\}$ in the column heads under Distinct and MS-Jaccard refer to $n$-gram. (\textbf{Bold}: the best; \underline{Underline}: the second best).}
	\label{tab-LM-results}
\end{table*}

\noindent\textbf{Dataset}\quad We performed experiments on the Wikitext-$103$\footnote{\url{https://s3.amazonaws.com/research.metamind.io/wikitext/wikitext-103-v1.zip}} dataset \cite{WIKITEXT-103} for language modeling. In order to train our POS Guided Softmax, we need the corresponding POS tags. We use the Stanford CoreNLP's POS tagger \cite{StanfordCoreNLP} to annotate words in Wikitext-$103$ with XPOS\footnote{The XPOS tags are language-specific part-of-speech tags from the Universal Dependency Treebanks.} tags \cite{CoNLL17}. In our implementation, there are 45 different POS tags in total.  

\noindent\textbf{Generation Details} \quad  We conduct the text completion task to evaluate models on the test set. Specifically, for each sample, we truncate 50 tokens as the prefix, and then guide model to decode following 100 tokens as the continuation from the given prefix. Finally, there are 1536 prefixes in the test set. We use stochastic decoding to generate text. Note that all the baselines have only one sampling stage, i.e., token sampling, while our POSG has an additional POS sampling. To reach a good trade-off between quality and diversity, we adopt nucleus sampling with $\alpha^{\text{(token)}} = 0.5$ for token sampling (for all models including our POSG and baselines). For our POSG, we adopt top-$k$ sampling in POS sampling, since the size of the POS vocabulary $\mathcal{P}$ is much smaller than the total token vocabulary. We then conduct a grid search to find the $k^{\text{(POS)}}$ whose generated continuations have the smallest reverse language model score~\cite{semeniuta2018accurate} on the validation set. $k^{\text{(POS)}}$ is finally set to $20$.  Some generated cases are shown in Appendix~\ref{case_study}.

\begin{table}[t]
    \centering
    \resizebox{\columnwidth}{!}{
			\begin{tabular}{l|cccccc}
				\toprule
				\multirow{2}{*}{Models} & \multicolumn{6}{c}{Distinct $\uparrow$}  \\ \cline{2-7} 
				& \multicolumn{1}{c}{1-P} & \multicolumn{1}{c|}{1-G} & 2-P & \multicolumn{1}{c|}{2-G} & 3-P       & 3-G     \\ \hline
		MLE & 16.0   & \multicolumn{1}{c|}{39.2}   & 38.5  & \multicolumn{1}{c|}{61.0}   & 54.7       & 70.8        \\
				FACE& 17.8   & \multicolumn{1}{c|}{49.8}   & 46.6  & \multicolumn{1}{c|}{73.0}   & 65.4        & 80.8        \\
				F$^2$-Softmax              & 16.4   & \multicolumn{1}{c|}{41.1}   & 39.3  & \multicolumn{1}{c|}{63.8}   & 55.8       & 74.0       \\
				UL  & 18.0   & \multicolumn{1}{c|}{51.7}   & 46.5  & \multicolumn{1}{c|}{76.8}   & 66.3       & 85.2      \\
				MoS &  16.4   & \multicolumn{1}{c|}{41.1}   & 40.0  & \multicolumn{1}{c|}{63.8}   & 56.5      & 72.9 \\
				POSG             & 19.9  & \multicolumn{1}{c|}{56.2}   & 58.1  & \multicolumn{1}{c|}{85.3}   & 80.6        & 92.3       \\
				Human                   & 21.7   & \multicolumn{1}{c|}{67.7}   & 61.8  & \multicolumn{1}{c|}{93.0}   & 83.8        & 95.9        \\ \hline
				PPMCC                   & \multicolumn{2}{c|}{0.988}           & \multicolumn{2}{c|}{0.986}          &
				\multicolumn{2}{c}{0.986} \\
				\bottomrule
		\end{tabular}}
	\caption{Results of distinct $n$-gram and $n$-POS with corresponding Pearson product-moment correlation coefficient (PPMCC). $n$-P and $n$-G where $n \in \{1,2,3\}$ are abbreviated notations for $n$-POS and $n$-gram.}
	\label{tab-LM-ppmcc}
\vskip -0.1in
\end{table}

\noindent\textbf{Metrics}\quad Following \citet{F2-Softmax}, we evaluate the generated text with two sets of metrics:  (i) \textbf{Diversity}: We use Self-BLEU \cite{self-bleu} which is calculated by computing BLEU \cite{bleu} of each generated text with all other generations as references. We also compute the generated continuations' unique tokens (Uniq), distinct $n$-gram (Distinct-$n$). We also use repetition (Rep) \cite{topp}, the percentage of continuations ending with a repetition loop, to evaluate text diversity. (ii) \textbf{Quality}: We measure the perplexity (PPL) \cite{perplexity}, KL-Divergence (KLD) \cite{KLDiver} on unigram distributions, and MS-Jaccard \cite{alihosseini2019jointly} on $n$-gram. All the metrics are calculated between the generations as hypotheses and the ground truths as references.

\noindent\textbf{Automatic evaluation}\quad
 Table~\ref{tab-LM-results} shows the automatic evaluation results comparing different models on the language modeling task. In terms of Self-BLEU$4$, Rep, and Distinct-$n$, our POSG performs much better than all the baselines, indicating that our proposed model can generate diverse text effectively. The FACE also performs well, and it achieves the best in Uniq. However, by checking the outputs (Table~\ref{case-LM} in Appendix~\ref{case_study}), we find that FACE produces more incoherent text that is hard to understand.

 Since training-based methods including ours make a trade-off between the text diversity and the likelihood of ground truth, MLE gets the lowest PPL. However, the optimal or second best results of quality metrics confirm that POSG can still maintain comparable generation quality.

 We further conduct a correlation test to verify that the text diversity is closely correlated with the POS diversity. We first randomly sample $500$ generated continuations from each model, and annotate them with the POS tagger. We define a $n$-POS to be contiguous $n$ POS tags from the annotated POS tag sequence. Then, we can describe the degree of POS diversity by calculating the proportion of the distinct $n$-POS. Table~\ref{tab-LM-ppmcc} presents results of distinct $n$-gram and $n$-POS with corresponding Pearson product-moment correlation coefficient. In terms of distinct $n$-POS, POSG also surpasses all the baselines. This demonstrates that our proposed model can substantially promote the POS diversity. Moreover, the Pearson correlations between distinct $n$-POS and $n$-gram are extremely high, which indicates that the high POS diversity indeed leads to the high text diversity.

\begin{table}[t]
\begin{center}
\begin{tabular}{l|ll}
\toprule
Models      & Div.  $\uparrow$       & Qua. $\uparrow$  \\ \cline{1-3}
	MLE     & 2.86$^{\star}$         &  3.10            \\
	FACE    &  3.32$^{\star}$        &  3.18              \\
	F$^2$-Softmax &  2.35$^{\star}$  &  2.80$^{\star}$         \\ 
	UL     &  3.36$^{\star}$         &  \textbf{3.20}    \\  
	MoS    &  2.79$^{\star}$         &  3.06$^{\star}$     \\ \hline
	POSG   &  \textbf{3.45}          &  3.17    \\
\bottomrule
\end{tabular}
\end{center}
\caption{Human evaluation on language modeling. $^{\star}$  denotes statistical significance compared with POSG (Mann-Whitney  $u$-test, $p <0.1$). }
\label{tab-HE-LM}
\vskip -0.1in
\end{table}

\noindent\textbf{Human evaluation}\quad For the language modeling task, following \citet{diversity-eval} we randomly sample $100$ generated continuations from each model. Each of them is scored between $1$ to $5$ ($5$ is the best), by five workers to evaluate the overall \textit{Diversity} (Div.) and \textit{Quality} (Qua.). The results of the human evaluation on language modeling are shown in Table~\ref{tab-HE-LM}. It can be seen that our POSG significantly outperforms all other baselines in diversity, and performs relatively well in quality.


\begin{table*}[t]
\centering
	\resizebox{\textwidth}{!}{
\begin{tabular}{l|ccccc|ccccc}
\toprule
\multirow{2}{*}{Models}   & \multirow{2}{*}{Self-WER$\uparrow$} & \multirow{2}{*}{Self-BLEU4$\downarrow$} & \multicolumn{3}{c|}{Distinct$\uparrow$}        & \multirow{2}{*}{BERTScore$\uparrow$} & \multirow{2}{*}{BLEU4$\uparrow$} & \multicolumn{3}{c}{ROUGE$\uparrow$}                                                                                    \\
                          &                     &                             & n=1                      & n=2           & n=3                        &                  &                        & 1        & 2             & L                    \\ \hline
MLE                     & 74.2                      & 25.1                        & 78.4                     & 82.8          & 78.5                 & 47.4                       & 9.81                   & 38.1          & 16.9          & 38.8                       \\
FACE                    & 73.0                      & 25.0                        & 78.9                     & \underline{83.6}          & 79.6      & 48.1                       & \underline{10.1}                   & 38.7          & 17.2          & 39.1                             \\
F$2$-Softmax             & 76.4                      & 28.0                        & 78.2 & 83.0          & 79.5        & \textbf{53.9}              & \textbf{11.4}          & \underline{41.1}    & \textbf{19.5} & \textbf{42.8} \\
UL                      & 77.2                      & \underline{21.2}                  & 80.1                     & \textbf{85.3}   & \underline{80.9}         & 36.0                       & 7.59                   & 30.6          & 13.3          & 30.3                         \\
SGCP                  & \underline{83.0}                & 28.6                        & \underline{81.9}               & 82.6          & 77.7               & 47.9                      & 9.91                   & \textbf{41.3} & \underline{17.7}    & \underline{41.1}               \\ \hline
POSG             & \textbf{89.7}             & \textbf{19.6}               & \textbf{82.1}            & \textbf{85.3} & \textbf{81.8}    & \underline{48.3}         &     9.79              & 40.3        & 17.1          &  39.4       \\
\bottomrule
\end{tabular}	}
		\caption{Automatic evaluation results for different models on the paraphrase generation task. Numbers $n \in \{1, 2, 3\}$ in the column heads under Distinct refer to $n$-gram. (\textbf{Bold}: the best; \underline{Underline}: the second best).}
	\label{tab-PARA-results}
\end{table*}

\begin{figure}[t]
	\begin{center}
		\centerline{\includegraphics[width=0.9\columnwidth]{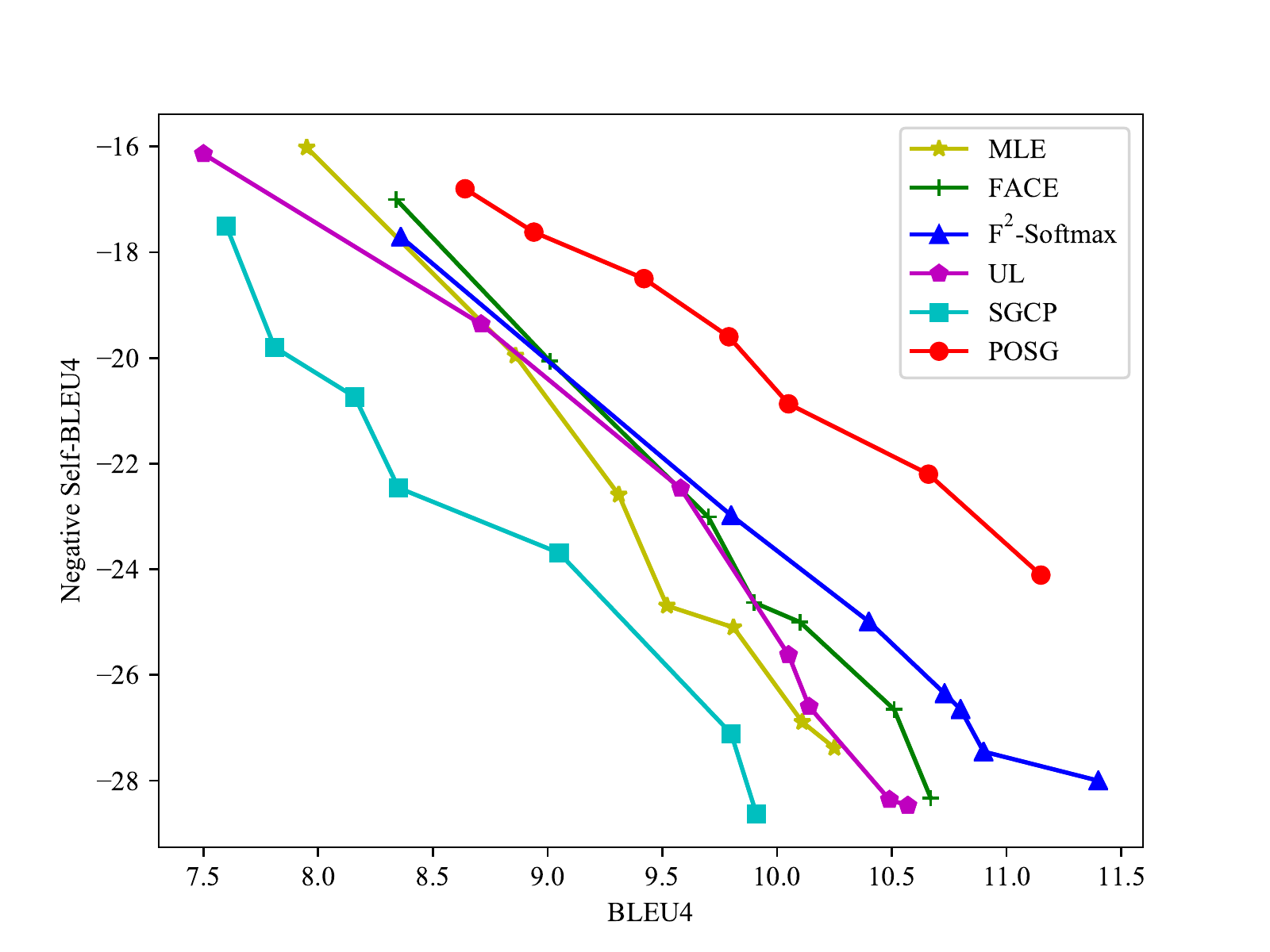}}
		\caption{Quality-diversity trade-off for different models on paraphrase generation. The x-axis measures BLEU4 for quality, and the y-axis measures negative Self-BLEU4 for diversity. Both are the bigger the better.}
		\label{fig-Q-D}
		\vskip -0.3in
	\end{center}
\end{figure}

\subsection{Paraphrase Generation}
\label{exper-para}

\noindent\textbf{Dataset}\quad We use the the ParaNMT-$50$M\footnote{\url{https://drive.google.com/file/d/1rbF3daJjCsa1-fu2GANeJd2FBXos1ugD/view}} dataset \cite{PARANMT50M} for paraphrase generation. ParaNMT-$50$M consists of over $50$ million paraphrases, generated by back-translation. For better training, we first remove the sentences that are less than $10$ tokens. Moreover, ParaNMT-$50$M dataset has provided translation scores to measure the quality of back-translation, that a low translation score means semantically inconsistent, while a high translation score usually accompanies low diversity. Therefore, we only keep the paraphrase pairs whose translation scores are between $0.7$ to $0.8$. Finally, we get a filtered dataset containing $1.6$ million paraphrase pairs with both high quality and diversity. We also use Stanford CoreNLP to tokenize the text and get corresponding POS tags.

\noindent\textbf{Generation Details}\quad We conduct the standard sequence-to-sequence paraphrase generation for testing. Note that, during inference, SGCP needs a corresponding exemplar sentence to paraphrase the input sentence, while our model does not. So, for a fair comparison, we prune the exemplar tree to the height $\max(3, H_{\text{max}} - 4)$ to reduce the impact from exemplar sentence, where $H_{\text{max}}$ is the height of the full constituency tree of the exemplar sentence. We use the test set provided in the work of SGCP\footnote{\url{https://github.com/malllabiisc/SGCP}} that contains $800$ paraphrase pairs and correspond exemplar sentences for inference. For a fair comparison, we closely follow \citet{SGCP} to generate paraphrase using beam search for all the models with beam size $10$. For the sampling hyperparameter in POS sampling, we also conduct a grid search, and $k^{\text{(POS)}}$ is finally set to $5$. Some generated cases are shown in Appendix~\ref{case_study}.

\begin{table}[t]
\begin{center}
\resizebox{\columnwidth}{!}{
\begin{tabular}{l|ll|ll}
\toprule
\multirow{2}{*}{Models}  & \multicolumn{2}{c|}{Div. $\uparrow$}   & \multirow{2}{*}{Flu. $\uparrow$} & \multirow{2}{*}{Rel. $\uparrow$}  \\
       & Lex.   & Syn.  &      &      \\ \hline
MLE    &   2.92 &   2.65$^{\star}$   &   3.34$^{\star}$   &   3.09$^{\star}$ \\
FACE   &   2.91&   2.58$^{\star}$    &   \textbf{3.60}&   3.35\\
F$^2$-Softmax &   2.77$^{\star}$    & 2.57$^{\star}$   &   3.59 &   \textbf{3.38} \\
UL    &   3.00 & 2.68$^{\star}$      & 3.37$^{\star}$  &   3.17$^{\star}$  \\
SGCP  &   2.74$^{\star}$ & 2.67$^{\star}$   &     3.50  &   3.21$^{\star}$ \\  
\hline
POSG  & \textbf{3.02} & \textbf{2.79}   &   3.58 &   3.35 \\
\bottomrule
\end{tabular}}
\end{center}
\caption{Human evaluation on paraphrase generation. $^{\star}$ denotes statistical significance compared with POSG (Mann-Whitney $u$-test, $p <0.1$).}
\label{tab-HE-PARA}
\end{table}

\begin{table*}[t]
	\centering
	\resizebox{\textwidth}{!}{
		\begin{tabular}{l|cccccc|ccccc}
			\toprule
			\multirow{2}{*}{Models} & \multirow{2}{*}{Self-BLEU4 $\downarrow$} & \multirow{2}{*}{Rep $\downarrow$} & \multirow{2}{*}{Uniq $\uparrow$} & \multicolumn{3}{c|}{Distinct $\uparrow$}  & \multirow{2}{*}{PPL $\downarrow$} & \multirow{2}{*}{KLD $\downarrow$} & \multicolumn{3}{c}{MS-Jaccard $\uparrow$}               \\
			 &   & & & n=1   & n=2    & n=3    &  &   & n=1 & n=2  & n=3   \\ \hline
			POSG   & \textbf{34.1}  & \textbf{0.000}        & \textbf{13.8k} & \textbf{60.2}        & \textbf{88.8}        & \textbf{94.3}     & 34.4  & \textbf{1.17}       & \textbf{62.2}        & \textbf{40.7}        & \textbf{25.9}     \\ 
			w/o POSG-Sampling    & 40.6  &  0.841       & 13.1k &  55.2     &  83.0       & 90.6     & 34.4    & 1.29      & 56.9        & 37.5        & 24.5    \\ 
		MLE  & 46.9 & 1.86  & 11.7k & 50.2 & 77.2 & 86.2    & \textbf{32.7} & 1.34   & 56.9  & 38.2 & 25.4  \\
\bottomrule
		\end{tabular}
	}
\caption{Results of ablation study on the language modeling task. Note that PPL measures the ability of the model to generate fluent text, which is not affected by the sampling strategy.}
	\label{tab-LM-ablation}
\end{table*}
\begin{table*}[t]
\resizebox{\textwidth}{!}{
\begin{tabular}{l|ccccc|ccccc}
\toprule
\multirow{2}{*}{Models}     & \multirow{2}{*}{Self-WER$\uparrow$} & \multirow{2}{*}{Self-BLEU4$\downarrow$} & \multicolumn{3}{c|}{Distinct$\uparrow$}      & \multirow{2}{*}{BERTScore$\uparrow$} & \multirow{2}{*}{BLEU4$\uparrow$} & \multicolumn{3}{c}{ROUGE$\uparrow$}                                                 \\
                              &                           &                             & n=1                      & n=2           & n=3            &                            &                        & 1        & 2             & L                        \\ \hline
POSG                & \textbf{89.7}             & \textbf{19.6}               & \textbf{82.1}            & \textbf{85.3} & \textbf{81.8}      & 48.3         &     9.79              & 40.3        & 17.1          &  39.4    \\   
w/o POSG-Sampling      &  87.6                       & 24.1                         & 78.0          & 80.5          & 78.5     &    \textbf{52.6}          & \textbf{11.1}           &\textbf{40.9} & \textbf{19.7} & \textbf{42.4}   \\
MLE         & 74.2                       & 25.1                         & 78.4          & 82.8          & 78.5      & 47.4                        & 9.81                    & 38.1                                                                                          & 16.9          & 38.8    \\
\bottomrule
\end{tabular}}
\caption{Results of ablation study on the paraphrase generation task. }
	\label{tab-PARA-ablation}
\end{table*}

\noindent\textbf{Metrics}\quad We also evaluate the generated paraphrases with two sets of metrics, (i) \textbf{Diversity}: To assess how different the generated paraphrases are compared to the original sentences, we calculate BLEU and Word Error Rate (WER) \cite{WER-paraphrase} between generated paraphrases and input sentences. We denote them as Self-BLEU (see Appendix~\ref{self-bleu-explanation} for the difference with the Self-BLEU in language modeling) and Self-WER, respectively. We also compute the generated paraphrases' distinct $n$-gram (Distinct-$n$) to evaluate text diversity. (ii) \textbf{Quality}: we calculate BLEU score on $n$-gram to evaluate the closeness of the generated paraphrases to references. Besides, we use the BERTScore \cite{BERTScore} to measure the semantic consistency between generated paraphrases and input sentences. We also compute ROUGE-$1$,$2$,L between the generated and the reference to evaluate the generation quality.

\begin{table}[t]
\resizebox{\columnwidth}{!}{
\begin{tabular}{l|ccc}
\toprule
Adjective   & Adjs. per            & \multirow{2}{*}{Self-BLEU4$\downarrow$}         & \multirow{2}{*}{BLEU4$\uparrow$} \\
Probability & \multicolumn{1}{l}{Sentence} &                        &                             \\ \hline
$\times 0.1$        &           0.43                    &        20.2               &         9.47                                \\ 
                      
$\times 1$          &               0.66               &       19.6                 &        9.79                               \\ 
$\times 10$        &             1.04                  &   18.7                     &         9.45                \\
\bottomrule
\end{tabular}}
\caption{Results of controllability analysis on the paraphrase generation task. ``$\times n$'' means that we manually multiply the probability of ``Adjective'' by $n$. }
	\label{tab-Controllability}
\end{table}

\noindent\textbf{Automatic evaluation}\quad The experimental results on the paraphrase generation task are shown in Table~\ref{tab-PARA-results}. Our proposed model outperforms other baselines on all the diversity metrics. In terms of quality metrics, our POSG performs better than MLE, FACE, and UL, while the best model in quality, i.e., F$^2$-Softmax performs badly in diversity. Moreover, compared with other syntax-guided models, i.e., SGCP, our model performs much better in diversity and has a comparable performance in quality. This further confirms that our model can effectively  promote text diversity without the help of exemplars.

To make a more intuitive comparison, we further apply stochastic decoding for different models, and tune the sampling hyper-parameters to generate different sets of paraphrases. Then, we calculate BLEU4 and Self-BLEU4 scores for these sets, and draw the quality-diversity trade-off in Figure~\ref{fig-Q-D}. Clearly, POSG surpasses all the baselines with a significant gap. These results confirm that our model can produce equally high-quality text that is more diverse, and vice versa.

\noindent\textbf{Human Evaluation}\quad We also conduct a human evaluation for the generated paraphrases. $100$ examples are randomly sampled from each models' outputs, respectively. Each of them are evaluated by five workers from the following four aspects: \textit{Lexical Diversity} (LeD.), and \textit{Syntactical Diversity} (SyD.), \textit{Fluency} (Flu.), \textit{Relevance} (Rel.). All these aspects are scored between 1 to 5, the higher the better. As shown in Table~\ref{tab-HE-PARA}, the results of the human evaluation are strongly consistent with the automatic evaluation. Compared with MLE, UL and SGCP, POSG substantially improves the generation quality, and it only has a tiny gap from the best model in fluency and relevance scores. Meanwhile, POSG has the best scores in diversity, which further verifies that our proposed methods can generate more lexically and syntactically diverse paraphrases. The detailed questionnaire, and other details are shown in Appendix~\ref{HEQuestion}.

\subsection{Ablation Study}

We perform ablation studies to reveal the effect of POS Guided Softmax and POS Guided Sampling. As shown in Table~\ref{tab-LM-ablation} and Table~\ref{tab-PARA-ablation}, compared with MLE, POS Guided Softmax (without POS Guided Sampling) can improve text quality for both the tasks. It is worth to mention that, it is natural to find that the model without POSG-Sampling performs better than the model with POSG. That is because POSG-sampling is a stochastic decoding method like nucleus sampling, which will sacrifice the quality of the generated text to promote text diversity. Therefore, POS Guided Sampling can dramatically promote text diversity for both the tasks. These results confirm the effectiveness of both the components. 

\section{Analysis}

\subsection{Interpretability}

Compared with one-stage sampling such as top-$k$ sampling, POSG will lead to the entropy increasing of a language model's distribution, and thus lead to more diverse outputs (see Appendix~\ref{proof} for the proof, Appendix~\ref{compared_one_stage} for experimental results).

\subsection{Controllability}
Our proposed POSG first samples a POS, and then samples a token from the vocabulary of the previously predicted POS. Therefore, we can control the POS sampling stage by forcing the probability of some specific POS to be higher or lower. For example, on the paraphrase generation task, we can multiply the probability of ``Adjective'' (``JJ'') and renormalize by dividing by the sum, aiming at generating more descriptive style paraphrases.

The results are shown in Table~\ref{tab-Controllability}. These results confirm that by leveraging POS as an observed and controllable clue, the generated text can be successfully modulated with negligible effect on quality and diversity  (see Appendix~\ref{controllability_analysis} for cases).

\section{Conclusion}

In this paper, we have introduced POS Guided Softmax and Sampling, simple but effective methods to address the low-diversity problem in text generation. POSG guides models to capture contextual and syntactical information by leveraging POS as an observed and controllable clue in both the training and decoding phases. Experimental results and human evaluation on language modeling and paraphrase generation have demonstrated the effectiveness of our methods.

\section*{Acknowledgements}
This work was supported by National Key R\&D Program of China (2021YFF0901502), National Science Foundation of China (No. 62161160339), State Key Laboratory of Media Convergence Production Technology and Systems and Key Laboratory of Science, Technology and Standard in Press Industry (Key Laboratory of Intelligent Press Media Technology). We appreciate the anonymous reviewers for their helpful comments. Xiaojun Wan is the corresponding author.

\bibliography{anthology,custom}
\bibliographystyle{acl_natbib}

\appendix

\section{Experimental Setup}
\label{DetailSetup}
\subsection{Dataset}
The dataset statistics of Wikitext-$103$ and  ParaNMT-$50$M are reported in Table~\ref{Stat-Wiki} and Table~\ref{Stat-Para}, respectively. 

Since ParaNMT-$50$M is generated by back-translation, the dataset has provided translation scores to measure the quality of back-translation, that a low translation score means semantically inconsistent, while a high translation score usually accompanies low diversity. Therefore, we only keep the paraphrase pairs whose translation scores are between $0.7$ and $0.8$. Moreover, for better training, we remove the sentences that are less than $10$ tokens. Finally, we get a filtered dataset containing $1.6$ million paraphrase pairs with both high quality and diversity. 

For language modeling, we use the original settings of Wikitext-$103$ dataset for training, validation, and test set splitting. For paraphrase generation, we use the filtered training, validation set of ParaNMT-$50$M, and the test set provided in the work of SGCP. It is worth to mention that Wikitext-$103$ is under the CC BY-SA $3.0$ license, and ParaNMT-$50$M is under the CC-BY license.

\begin{table}[h]
\begin{center}
\begin{tabular}{c|ccc}
\toprule
           & Train       & Valid   & Test    \\ \hline
\#Articles & 28,475      & 60      & 60      \\
\#Tokens   & 103,227,021 & 217,646 & 245,569 \\
\bottomrule
\end{tabular}
\end{center}
\caption{Statistics of Wikitext-$103$.}
\label{Stat-Wiki}
\end{table}

\begin{table}[h]
\begin{center}
\begin{tabular}{c|ccc}
\toprule
           & Train       & Valid   & Test    \\ \hline
\#Sentence & 1,640,709      & 3,000      & 800      \\
\bottomrule
\end{tabular}
\end{center}
\caption{Statistics of ParaNMT-$50$M.}
\label{Stat-Para}
\end{table}

\subsection{Architectures and Hyperparamters}

For the language modeling task, we use a $12$-layer Transformer Decoder with $8$ attention heads, embedding dimension $512$, and projection dimension $2048$. For the paraphrase generation task, we use a $6$-layer Transformer Encoder and Decoder with the same other settings. All the algorithms are implemented in Pytorch and trained on a machine with 8 NVIDIA GTX 2080Ti GPUs for $10$ epochs with the hyper-parameters reported in Table~\ref{hyper-parameters}. 

\begin{table}[h]
\small
\begin{center}
\begin{tabular}{l |c | c}
\toprule
 hyper-parameters & Wikitext-$103$ & ParaNMT-$50$M \\
\hline 
Vocabulary size & 267,735 & 100,000 \\
 Batch size & 12 & 96 \\
 Learning rate & 0.0001 & 0.0001\\
 Finetuning LR & 0.00001 & 0.00001\\
 Finetuning step & 1500 & 1500 \\
 Gradient clipping & 0.25 & 0.25 \\
 Weight decay & 0.001 & 0.001\\
 Droupout & 0.1 & 0.1 \\
 Optimizer  & \text { Adam } & \text { Adam }\\
\quad  -$\beta_{1}$& 0.9 & 0.9  \\
\quad  -$\beta_{2}$  & 0.999 & 0.999 \\
\quad  -$\epsilon$ & 1$\mathrm{e}$-8 & 1$\mathrm{e}$-8\\
\bottomrule
\end{tabular}
\end{center}
\caption{Hyperparameter settings for different datasets.}
\label{hyper-parameters} 
\end{table}

We choose the architecture settings and batch sizes according to the GPU memory constraint. Note that we use FACE-OPR among the four variants of FACE, and we train it in the way of finetuning with corresponding finetuning LR and finetuning step. Additionally, we use $7$ mixture components in MoS.

\subsection{Metrics}
\label{self-bleu-explanation}
Note that, the calculations of Self-BLEU are different for language modeling and paraphrase generation. This is because the typical definitions of Self-BLEU for these two different task are indeed different. For language modeling, Self-BLEU~\cite{self-bleu} is a metric to evaluate the inner diversity of the generated data, while for paraphrase generation, Self-BLEU~\cite{DivGAN} is used to evaluate the degree to which the generated paraphrases are different from the original sentence.

\section{Proof}
\label{proof}
We prove that our POS Guided Softmax and Sampling can certainly generate more diverse text than the one-stage sampling, top-$k$ sampling as an example. 

In information theory, the entropy of a random variable is the average level of ``information'', ``surprise'' in the variable's possible outcomes. Therefore, we can use the entropy of a language model's distribution $p(x)$ to measure its diversity. We denote the entropy of $p(x)$ as $H(p)$: $H(p) = -\sum_{x\in \mathcal{V}}p(x)\log p(x)$. The increase of $H(p)$ means the increase of diversity.

For example, compared with greedy search, diversity-promoting sampling methods, such as top-$k$ sampling can increases $H(p)$ from $ -p(x_{\text{max}})\log p(x_{\text{max}})$ to $-\sum_{x\in \mathcal{V}_{k}}\frac{p(x)}{Z_k}\log \frac{p(x)}{Z_k}$, where $x_{\text{max}}$ is the token with the max probability, $\mathcal{V}_{k}$ is the set of top-$k$ most probable tokens, $Z_k = \sum_{x\in \mathcal{V}_{k}}p(x)$, and obviously $x_{\text{max}} \in \mathcal{V}_{k}$.

Now, we prove that our POSG with two sampling stages can lead to the entropy increasing, compared with one-stage top-$k$ sampling as an example. 

For one-stage top-$k$ sampling,

\begin{equation}
\small
    \begin{aligned}
		\label{eq-topk-1}
	&	H \left(p \right)^{\text{(top-k)}} = -\sum_{x\in \mathcal{V}_{k}}\frac{p(x)}{Z_k}\log \frac{p(x)}{Z_k} \\
		&= -\sum_{x\in \mathcal{V}_{k}}\frac{\sum_{\rho \in \mathcal{P}} p(x, \rho)}{Z_k}\log \frac{\sum_{\rho \in \mathcal{P}} p(x, \rho)}{Z_k} \\
&= -\log |\mathcal{P}| -\sum_{x\in \mathcal{V}_{k}}\frac{\sum_{\rho \in \mathcal{P}} p(x, \rho)}{Z_k}\log \frac{\sum_{\rho \in \mathcal{P}} p(x, \rho)}{Z_k \times |\mathcal{P}|}
    \end{aligned}
\end{equation}
According to the Log sum inequality, it follows:

\begin{equation}
\small
    \begin{aligned}
		\label{eq-topk-2}
   & H(p)^{\text{(top-k)}}  \ge  -\log |\mathcal{P}| -\sum_{x\in \mathcal{V}_{k}} \sum_{\rho \in \mathcal{P}}\frac{ p(x, \rho)}{Z_k}\log \frac{p(x, \rho)}{Z_k } \\
   & = -\log |\mathcal{P}| -\sum_{\rho \in \mathcal{P}} \sum_{x\in \mathcal{V}_{k}}  \frac{ p(x, \rho)}{Z_k}\log \frac{p(x, \rho)}{Z_k }  
    \end{aligned}
\end{equation}
Since $p (x, \rho) = 0$ for $x \notin \mathcal{V}_{\rho} $, it follows:

\begin{equation}
\small
    \begin{aligned}
		\label{eq-topk-3}
		H(p)^{\text{(top-k)}}  \ge  -\log |\mathcal{P}| -\sum_{\rho \in \mathcal{P}} \sum_{x\in \mathcal{V}_{k, \rho}}  \frac{ p(x, \rho)}{Z_k}\log \frac{p(x, \rho)}{Z_k } 
    \end{aligned}
\end{equation}

where $\mathcal{V}_{k, \rho} = \{x\in \mathcal{V}_k \mid \rho \in \text{POS}(x) \}$, $\text{POS}(x)$ is the set of all POS tags of token $x$. Thus, $\mathcal{V}_{k, \rho} \subseteq \mathcal{V}_{k} $.

For our POSG with two sampling stages,

\begin{equation}
\small
    \begin{aligned}
		\label{eq-poss-1}
	&	H(p)^{\text{(POS)}}  =  -\sum_{x\in \mathcal{V}} \sum_{\rho \in \mathcal{P}} p^\prime(x, \rho)\log  \sum_{\rho \in \mathcal{P}} p^\prime(x, \rho) \\
	&= -\sum_{x\in \mathcal{V}} \sum_{\rho \in \mathcal{P}} p^\prime(\rho) p^\prime(x \mid \rho ) \log  \sum_{\rho \in \mathcal{P}} p^\prime(\rho) p^\prime(x \mid \rho ) 
    \end{aligned}
\end{equation}

where $p^\prime(x, \rho)$ is defined in Equation~\ref{joint-p},  $p^\prime(\rho)$ and  $p^\prime(x \mid \rho )$ are defined in Equation~\ref{eq-posg-sampling}. Again, according to the Log sum inequality, it follows:

\begin{equation}
\small
    \begin{aligned}
    \label{eq-poss-2}
	 & H(p)^{\text{(POS)}} \ge -\log |\mathcal{P}|\\
	& -\sum_{x\in \mathcal{V}} \sum_{\rho \in \mathcal{P}} p^\prime(\rho) p^\prime(x \mid \rho ) \log   p^\prime(\rho) p^\prime(x \mid \rho )
    \end{aligned}
\end{equation}

\begin{table*}[t]
\resizebox{\textwidth}{!}{
\begin{tabular}{l|ccccc|ccccc}
\toprule
\multirow{2}{*}{Models} & \multirow{2}{*}{Self-WER$\uparrow$} & \multirow{2}{*}{Self-BLEU4$\downarrow$} & \multicolumn{3}{c|}{Distinct$\uparrow$}        &
\multirow{2}{*}{BERTScore$\uparrow$} & \multirow{2}{*}{BLEU4$\uparrow$} & \multicolumn{3}{c}{ROUGE$\uparrow$}                                                                                   \\
                         &                           &                             & n=1                      & n=2           & n=3                     &                            &                        & 1        & 2             & L              \\ \hline

top-$k$                 & 100.8                      & 13.6                    & 86.9        & 88.9         & 83.7      & 39.4                       & 6.49                   & 33.5                                                                                         & 12.1        & 32.3    \\  
POSG                & 102.1             & 13.6              & 86.9   & 88.2  & 83.4                    & \textbf{43.3}                      & \textbf{7.71}                 & \textbf{36.4}                                                                                      & \textbf{14.1}         & \textbf{34.9}              \\  \hline
\quad $\Delta$     & +1.3   & +0.0  & +0.0  & -0.7  &  -0.3  &  \textbf{+3.9}  &  \textbf{+1.22}  & \textbf{+2.9}  &  \textbf{+2.0}  & \textbf{+2.6} \\
\bottomrule
\end{tabular}}
\caption{Results of POSG and one-stage sampling (we use top-$k$ sampling here) on the paraphrase generation task. Note that we tune the sampling hyper-parameters of both methods to reach the same level of diversity, and then compare the text quality.}
	\label{tab-additional-sampling-ana}
\end{table*}

For the sake of briefness and fairness, we assume that our POSG adopts pure sampling in the first sampling stage (POS Sampling), and adopts top-$k$ sampling with the same $k$ in the second sampling stage (Token Sampling). So, $p^\prime(\rho) = p(\rho)$ for $\rho\in \mathcal{P}$ , while 

$$
\small
p^{\prime}(x \mid \rho) = \begin{cases}\frac{p(x \mid \rho)}{\mathcal{Z}_{2}}, & \text { if } x \in \mathcal{V}_{\rho, k} \\\\ 0, & \text { otherwise }\end{cases}, Z_2 = \sum_{x\in \mathcal{V}_{\rho, k}}p(x)
$$

Note that, in our paper, we denote all the tokens whose POS is $\rho$ as a vocabulary $\mathcal{V}_{\rho}$, and here, $\mathcal{V}_{\rho, k}$ is the set of top-$k$ most probable tokens in $\mathcal{V}_{\rho}$. Thus, $\mathcal{V}_{\rho, k} \subseteq \mathcal{V}_{\rho} $. Then, it follows:
\begin{equation}
\small
    \begin{aligned}
    \label{eq-poss-3}
    & H(p)^{\text{(POS)}} \ge -\log |\mathcal{P}|\\
    & -\sum_{x\in \mathcal{V}} \sum_{\rho \in \mathcal{P}} p(\rho) p^\prime(x \mid \rho ) \log   p(\rho) p^\prime(x \mid \rho ) \\
& = -\log |\mathcal{P}| -\sum_{\rho \in \mathcal{P}}\sum_{x\in \mathcal{V}_{\rho, k}}  p(\rho) \frac{p(x \mid \rho )}{Z_2} \log   p(\rho) \frac{p(x \mid \rho )}{Z_2} \\
& =  -\log |\mathcal{P}| -\sum_{\rho \in \mathcal{P}}\sum_{x\in \mathcal{V}_{\rho, k}} \frac{p(x, \rho )}{Z_2} \log \frac{p(x, \rho )}{Z_2}
    \end{aligned}
\end{equation}

Since $\mathcal{V}_{k, \rho} \subseteq  \mathcal{V}_{\rho, k}$ and we use the same setting of $k$, i.e., $Z_2 \approx Z_k$, we can finally conclude from Equation~\ref{eq-topk-3} and Equation~\ref{eq-poss-3} that the lower bound of $H(p)^{\text{(POS)}}$ is greater than or equal to the lower bound of $H(p)^{\text{(top-k)}}$. When compared with other one-stage sampling strategies, this conclusion still holds, and can be proved in a similar way. Consequently, this will account for the effectiveness of our methods.

\section{Additional Analysis}
\label{addition_ala}
\subsection{Compared with One-stage Sampling}
\label{compared_one_stage}
We further conduct an analysis to test whether the traditional one-stage sampling can achieve the same level of diversity by increasing the randomness, e.g. using larger $k$ in top-$k$ sampling. On the paraphrase generation task, we tune the sampling hyper-parameters in top-$k$ sampling and our POSG to reach the same level of diversity, and then compare the text quality. The results are shown in Table~\ref{tab-additional-sampling-ana}. In this experiment, POSG adopts top-$k$ sampling with $k^{(\text{POS})}=5$ in POS sampling,  $k^{\text{(token)}}=500$ in token sampling, while MLE adopts top-$k$ sampling with $k=1000$. Obviously, our POSG significantly outperforms top-$k$ sampling on MLE in terms of quality metrics, while performing equally well in diversity. Therefore, we can conclude that, by increasing the randomness, the traditional one-stage sampling on MLE can finally achieve the same level of diversity as our POSG, but the quality of the generated text will seriously deteriorate. This further confirms the advantage of our methods over prior works.
\subsection{Controllability Analysis Example}
\label{controllability_analysis}
An example of the controllability analysis is provided in Table~\ref{Example-Controllability}. When we control the probability of adjective increasing during the POS sampling stage, the generated paraphrase will contain correspondingly more adjectives.

\begin{table}[h]
	\begin{center}
	\resizebox{\columnwidth}{!}{
		\begin{tabular}{l p{0.97\columnwidth}}
			\toprule
	\textbf{Input Sentence}: &  he (PRP) was (VBD) smiling (VBG) , clearly (RB) \textbf{delighted (JJ)}\\\hline \hline
\textbf{$\times 0.1$} & he (PRP) was (VBD) smiling (VBG) , and (CC) he (PRP) was (VBD) clearly (RB) pleased (VBN) with (IN) joy (NN)\\\hline
\textbf{$\times 1$}  & he (PRP) was (VBD) smiling (VBG) and (CC) apparently (RB) \textbf{delighted (JJ)} with (IN) joy (NN) in (IN) his (PRP\$) face (NN) \\ \hline
\textbf{$\times 10$}  & he (PRP) was (VBD) still (RB) smiling (VBG) and (CC) \textbf{delighted (JJ)} with (IN) \textbf{apparent (JJ)} joy (NN) in (IN) his (PRP\$) face (NN)  \\ 
			\bottomrule
		\end{tabular}
	}
	\end{center}
		\caption{\label{Example-Controllability}Examples of controllability analysis on the paraphrase generation task.}
\end{table}

\subsection{Tuning $\alpha^{\text{(POS)}}$ and $k^{\text{(POS)}}$}

\begin{figure}[h]
	\begin{center}
		\centerline{\includegraphics[width=\columnwidth]{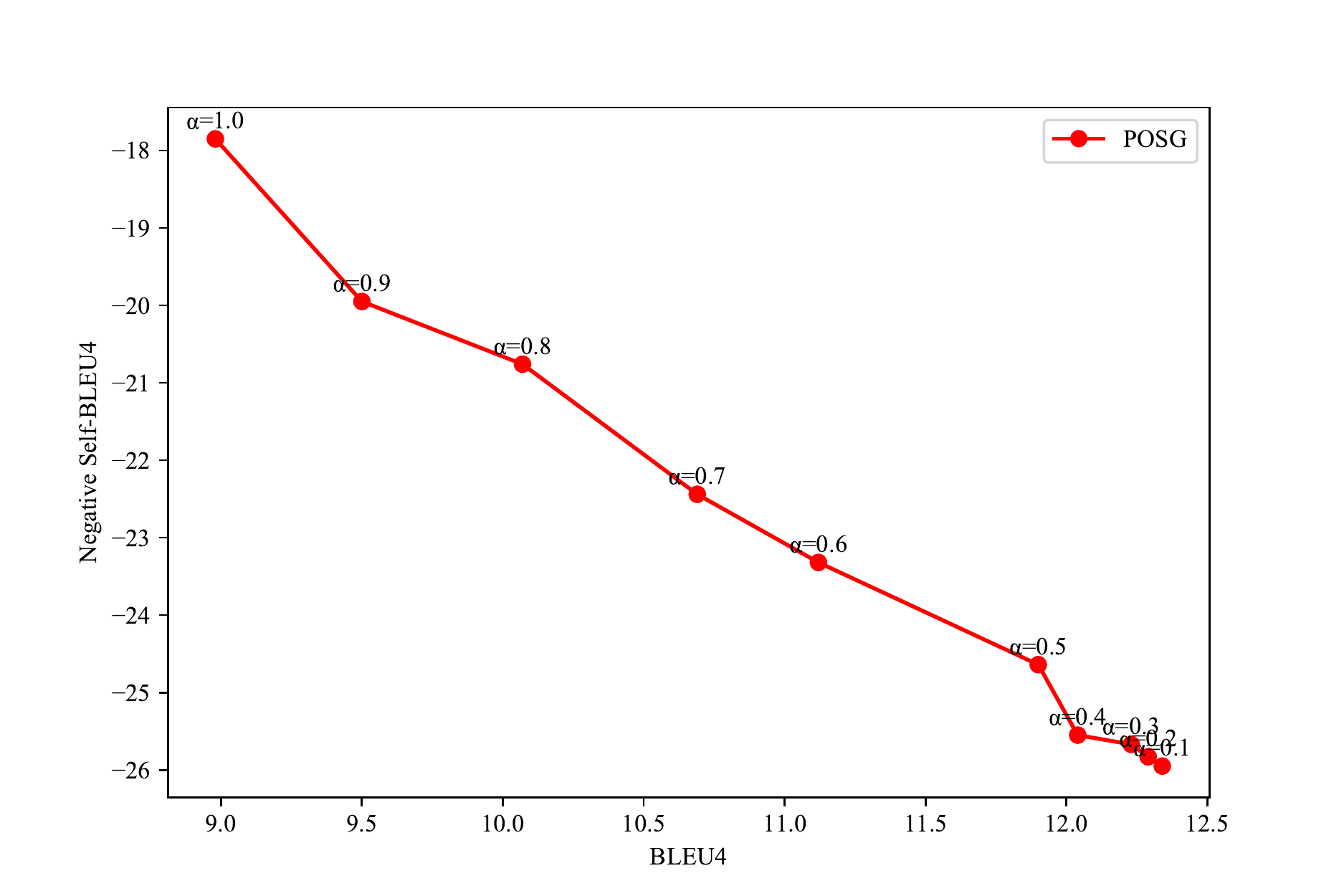}}
		\caption{Quality-diversity trade-off for POSG on paraphrase generation by tuning $\alpha^{\text{(POS)}}$.}
		\label{fig-Q-D-pos-topp}
		\vskip -0.2in
	\end{center}
\end{figure}
\begin{figure}[h]
	\begin{center}
		\centerline{\includegraphics[width=\columnwidth]{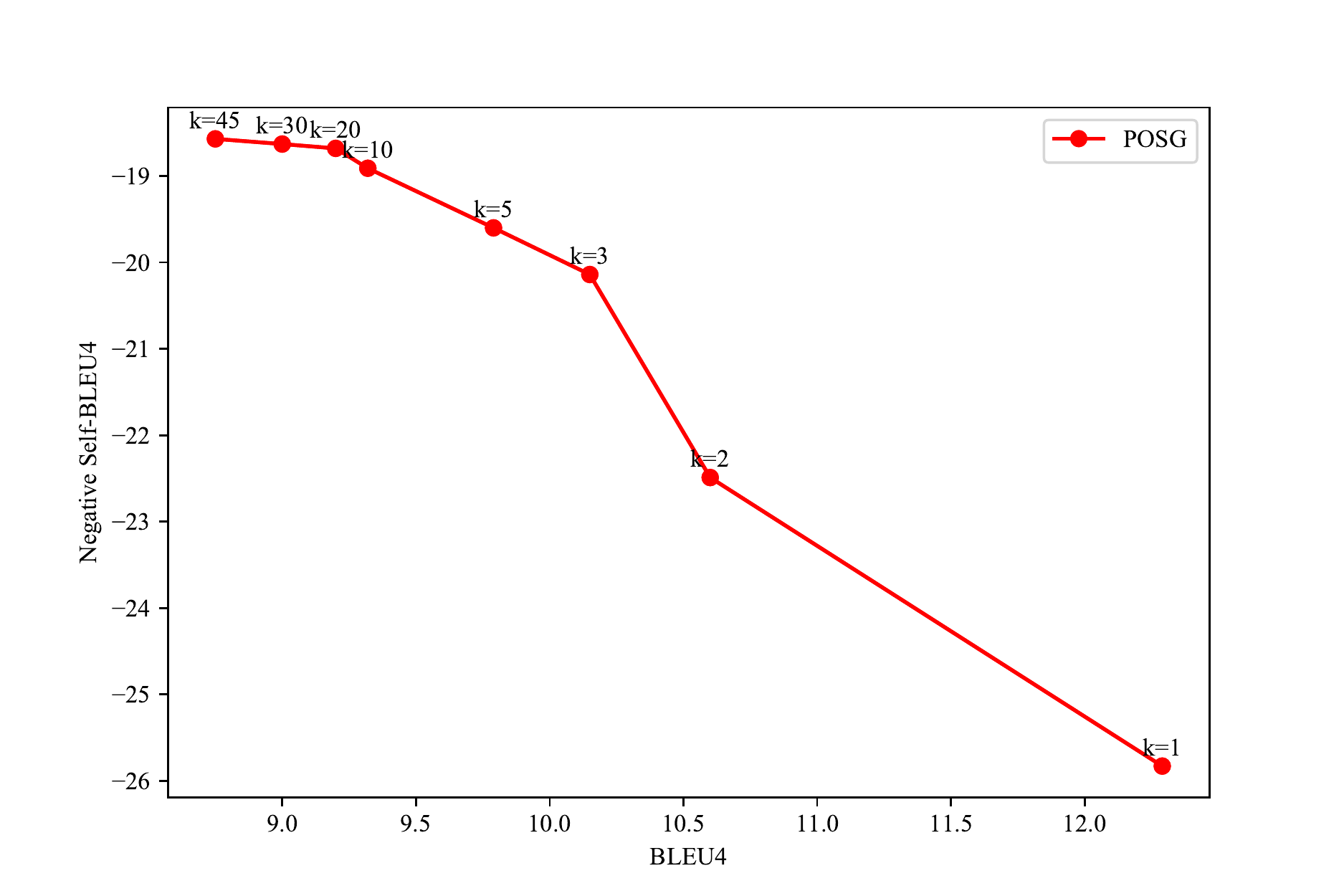}}
		\caption{Quality-diversity trade-off for POSG on paraphrase generation by tuning $k^{\text{(POS)}}$.}
		\label{fig-Q-D-pos-topk}
		\vskip -0.2in
	\end{center}
\end{figure}
We have conducted an additional analysis on paraphrase generation by tuning $\alpha^{\text{(POS)}}$ or $k^{\text{(POS)}}$ while fixing all other hyper-parameters. The results are shown in Figure~\ref{fig-Q-D-pos-topp} and Figure~\ref{fig-Q-D-pos-topk}. It can be observed that: with other sampling hyper-parameters fixed, as $\alpha^{\text{(POS)}}$ or $k^{\text{(POS)}}$ increases, the results of diversity metrics get better, but the results of quality metrics get worse.

\begin{table*}[t]
\small
	\centering
	\resizebox{\textwidth}{!}{
		\begin{tabular}{p{\textwidth}}
			\toprule
			\textbf{Prefix}: Below them, North Koreans continued crossing the river and moving supplies forward to their combat units, some of them already several miles eastward. The North Koreans quickly discovered Task Force Manchu group. They first attacked it at 14:00 that afternoon, and were repulsed \\
			\hline
			\hline
			\textbf{MLE}:   by the North Koreans. On the morning of September 8, \textcolor{orange}{the North Korean forces launched a} surprise \textcolor{orange}{attack} on the high ground west of the Kum River. At 16:30, \textcolor{orange}{the North Korean force launched a} massive \textcolor{orange}{attack} on the North Korean force, but the initial attack was not successful. The North Korean offensive was halted by the remnants of the North Korean forces  \\
			\hline
			\textbf{FACE}:  by heavy machine-gun fire. In the early morning hours of 8 September, North Korean troops \textcolor{red}{were alerted to attack on the perimeter}. On 9 September, a force of 20,000 men led by Lieutenant Colonel Robert E. \textcolor{red}{telluride} began to attack the North Korean lines, suffering little damage. By 14:00 on 9 September, North Korean forces had crossed the \textcolor{red}{Naktong} River just before midnight.  \\
			\hline
			\textbf{F$^2$-Softmax}:  by the North Koreans. The North Koreans were ordered to withdraw to the rear of the North Koreans. They \textcolor{orange}{then launched a frontal attack} on the south side of the river. The North Koreans \textcolor{orange}{then launched a frontal attack} on the North Korean right flank. The North Korean right flank was soon overrun by \textcolor{blue}{the North Koreans}. \textcolor{blue}{The North Koreans} were subsequently repulsed by \textcolor{blue}{the North Koreans}, \\
			\hline
			\textbf{UL}:   by North Korean fire, which forced the North Koreans to retreat. A further assault by the 1st U.S. Infantry Regiment followed in the afternoon, and after seven hours of fighting, the 2nd U.S. Infantry Regiment broke off the attack and retreated across the river. The survivors of the Battle \textcolor{red}{of tellers} managed to escape to a new bridge. \textcolor{red}{Task Force presaged}, but by 20:00 the North Koreans were completely surrounded by North Korean troops.\\\hline
			\textbf{MoS}:   by the 9th Infantry Regiment. At 17 : 00, the North Koreans took the road from the Korean border to the north, and began firing on the northern flank of the North Korean forces. The North Koreans then withdrew to the northern flank of the Korean army, where they advanced into the river and quickly attacked the North Koreans. At 16 : 00, \textcolor{blue}{the North Koreans} began firing on \textcolor{blue}{the North Koreans}, and a number of \textcolor{blue}{North Korean} soldiers, including the 5th Cavalry Regiment, \textcolor{red}{attacked} \textcolor{blue}{the North Koreans}.  \\
			\hline
			\textbf{POSG}: by the North Koreans, beginning their advance south of Osan on 18 September. By nightfall on 24 September, Ho Chi Minh had secured its flank, while the South Koreans had captured the town of Phong on the west of Taejon. The North Koreans had retreated to Pyongtaek, and in the afternoon of 22 September two North Koreans were killed there, leaving behind the town to the survivors.\\
			\bottomrule
		\end{tabular}
	}	\caption{Examples of language modeling on Wikitext-$103$ dataset. Repeating text is highlighted in \textcolor{blue}{blue}, dull text with single context is highlighted in \textcolor{orange}{orange}, and incoherent text is highlighted in \textcolor{red}{red}.}
	\label{case-LM}
	\vskip -0.1in
\end{table*}

\begin{table}[t]
	\small 
	\centering
		\begin{tabular}{p{0.95\columnwidth}}
			\toprule
			\textbf{Source}: this is going to make good economic sense for the city .  \\
			\textbf{Reference}: that it would be good for the city in a certain economic sense .  \\
			\hline
			\hline
			\textbf{MLE}:  this will be an economic sense for the entire city .  \\
			\hline
			\textbf{FACE}:  this will create a good economic point in the city . \\
			\hline
			\textbf{F$^2$-Softmax}:  this will make sense of economic sense for the city . \\
			\hline
			\textbf{UL}:  this will be considerable economic considerations for the city 's going to be able to economic point of the city . \\
			\hline
			\textbf{SGCP}:  this will make economic sense for the city .  \\
			\hline
			\textbf{POSG}:  it is what makes good economic sense to the city . \\
			\bottomrule
		\end{tabular}
		   \caption{Examples of paraphrase generation on ParaNMT-$50$M. }
	\label{case-PARA}
\end{table}

\section{Case Study}
\label{case_study}
Table~\ref{case-LM} provides examples of text completion produced by our model and other baselines. It can be observed that MLE, F$^2$-Softmax, and MoS suffer from a severe repetition problem, and they also generate many similar sentences about a single content. Due to the low-diversity problem, MoS even generates some illogical text, such as ``the North Koreans began firing on the North Koreans''.  FACE produces a large amount of incoherent text, making the text somewhat hard to read. UL and our POSG alleviate those problems, while our model performs relatively better.

Additionally, examples of paraphrase generation are shown in Table~\ref{case-PARA}. We observe that almost all models can generate high-quality paraphrases with well-preserved semantic meanings, while our POSG exhibits more syntactic diversity than other baselines.

\section{Human Evaluation}
\label{HEQuestion}

We post the human evaluation questionnaire, as shown in Table~\ref{Question-LM} and Table~\ref{Question-PARA}, and then recruit five workers with sufficient high English skills. We pay each worker 60 US dollars, and let them complete the evaluation within a week.

For both tasks, workers are given 100 randomly sampled inputs, and corresponding outputs from each model. Then, they need to score those outputs according to the description in the questionnaire. The term ``diversity'' in language modeling is typically regarded as a property of the collective outputs of a system, but it is really difficult for a human to remember such a large scale of outputs and give an overall score for a system. So we make a compromise that we asked the worker to rate the diversity of individual outputs, and intuitively the more diverse individual outputs are, the more diverse the system is.

We employ the Krippendorff's alpha for the inter-annotator agreement analysis. As shown in Table~\ref{tab-HE-LM-IAA} and Table~\ref{tab-HE-PARA-IAA}, all the results are fair agreement ($0.2 \le \kappa \le 0.4$) or moderate agreement ($0.4 \le \kappa \le 0.6$).

\begin{table}[h]
\begin{center}
\begin{tabular}{l|ll}
\toprule
		 & Div.   & Qua. \\  \hline
		Krippendorff's $\alpha$    &  0.57   &  0.40     \\
		\bottomrule
\end{tabular}
\end{center}
\caption{Agreement analysis for annotators labels on the language modeling task.}
\label{tab-HE-LM-IAA}
\end{table}
\begin{table}[h]
\begin{center}
\begin{tabular}{l|ll|ll}
\toprule
\multirow{2}{*}{Models}  & \multicolumn{2}{c|}{Div.}  & \multirow{2}{*}{Flu.} & \multirow{2}{*}{Rel.} \\
 & Lex.   & Syn.  &      &         \\ \hline
Krippendorff's $\alpha$  &  0.54  & 0.37   &   0.71  &  0.63  \\
\bottomrule
\end{tabular}
\end{center}
\caption{Agreement analysis for annotators labels on the paraphrase generation task.}
\label{tab-HE-PARA-IAA}
\end{table}

\section{Impact of the POS tagger}
\label{impact_tagger}
In our work, we use an off-the-shelf POS tagger to annotate the POS tags, and build the POSG upon these annotated POS tags. Consequently, the better the quality of POS tagging, the better the performance of our method. Stanford CoreNLP's POS tagger \cite{StanfordCoreNLP}, the POS tagger we use, is one of the state-of-the-art tagger, which is the most commonly used tool for NLP research. This ensures the high quality of tagging results.

\begin{table*}[t]
	\begin{center}
	\small
	\resizebox{\textwidth}{!}{
		\begin{tabular}{p{\textwidth}}
			\toprule
The goal of this review is to evaluate the quality and diversity of generated texts. In this review, you will read an excerpt from Wikipedia with first 50 words as prefixes, and its corresponding 100-word continuations. You should rate the continuations between 1 - 5 in two ways: \\
(1)	Diversity. The overall diversity of text can be evaluated from form (How to say it?) and content (What to say?). (1 = The continuation is always repeating some words, its sentences share the similar forms syntactically and lexically, and its content is dull; 5 = The continuation seldom repeats words, its sentences have various syntactical and lexical forms, and it contains different things related to the prefix)\\
(2)	Quality. The overall quality of text can be evaluated in many different aspects, such as fluency, readability, coherence, and so on. (1 = The continuation is incoherent, difficult to understand, not related to the prefix, and has many syntactically and semantically errors; 5 = The continuation is coherent, easy to understand, related to the prefix, and grammatically correct) \\
You should score between 1 - 5, where 5 is best and 1 is worst. You can consider and make a final decision by comparing different continuations of the same prefix. These prefixes and continuations have been preprocessed by separating punctuation, and splitting conjunctions. And because of length constraints, they may be truncated in the middle of the text. \textbf{Please ignore these when evaluating and do not allow them to affect your judgments.}  \\ \hline
For example, if you are given the following prefix and continuations: \\
\textbf{Prefix}:
Infantry fought a series of hard battles . Simultaneously it had to send combat patrols to its rear to clear infiltrating North Koreans from Changnyong and from its supply road . On the morning of September 1 the 1st and 2nd Regiments of the NK 9th Division , in their…

\textbf{Continuation 1}: \centerline{Score: Diversity 3, Quality 3} \\
second action around 380 yards \textcolor{red}{$<$unk$>$ 290 m $<$unk$>$ from .395} , were assigned to the 7th Division. Meanwhile , the 2nd Regiment of the 27th Infantry , which had been sent out of Murmansk in late October , had moved to \textcolor{red}{iconoclasts Pass} . It was the first major US unit to advance across the mountains to the west , but it was not successful . On the morning of September 2 , the Eighth Army's 2nd Battalion , 5th Cavalry Regiment was moved into a position about 10 miles $<$unk$>$ 16 km $<$unk$>$ from Highton .

\textbf{Continuation 2}:  \centerline{Score: Diversity 1, Quality 5} \\
reserve , were positioned to engage the North Korean troops . The 3rd Regiment of the NK 10th Division was to attack the North Korean forces . \textcolor{orange}{The 1st Regiment of the NK} 12th \textcolor{orange}{Division was to attack the North Koreans from the} north and east of the 2nd Division , and \textcolor{orange}{the 1st Regiment of the NK 15th Division was to attack the North Koreans from the} south . \textcolor{orange}{The 1st Regiment of the NK 12th Division was to attack the North Koreans from the south . The 1st Regiment of the NK 15th Division was to attack the}

\textbf{Continuation 3}: \centerline{Score: Diversity 1, Quality 1}\\
position north of Wonju , were repeatedly pushed back by the ROK 3rd Division . At 08 : 00 am the units of the 1st Battalion attempted to attack . Kim of the \textcolor{blue}{1st and 2nd Battalions attacked the 3rd and 3rd Battalions of the 2nd Battalion of the 3rd Battalion of the 3rd Battalions of the 1st Battalion of the 2nd Battalion of 2nd Battalion} , 7th Marines on North , 7th Marines on Hill 60 . Task Force 51 and 9th Marines attacked Sangju 's \textcolor{blue}{1st Battalion of the 3rd Battalion of the 2nd Battalion} , 1st Platoons

\textbf{Continuation 4}:  \centerline{Score: Diversity 4, Quality 3}
`` Series B '' Company , carried out three assaults on the Pusan on 29 September against three resistance groups that included the \textcolor{red}{blacksmiths} , truck commanders , and air support . They then conducted three raids into a line \textcolor{red}{south of psalmody} by the 2nd Battalion , 3rd Field Artillery Regiment . At the same time , units from the 3rd Infantry Division and the 3rd Marine Division advanced on all four sides of the road , while infantry units of the 2nd Infantry Division advanced on the northern slope . The 5th Marine Corps , in particular\\ \hline
\textbf{Analysis}:
As for diversity, Continuation 1 gives various details about the ``hard battles'', and is of high diversity in the text form. But all the content of it is about the deployment of armies, which means low content diversity. Therefore, Continuation 1 gets 3 points in Diversity. Since there are some words difficult to understand (highlighted in \textcolor{red}{red}), Continuation 1 gets 3 points in Quality.

Continuation 2 keeps talking about only one single content, that is ``some Regiment attacks the North Koreans from somewhere'' (highlighted in \textcolor{orange}{orange}). Although it is fluent, relevant, and gets high scores in Quality, Continuation 2 will receive the lowest score in Diversity due to its dull content.

Continuation 3 contains many useless repeating text (highlighted in \textcolor{blue}{blue}), which makes the continuation incoherent and hard to understand, so it gets the lowest score in both Quality and Diversity.

Continuation 4 also states from many different aspects of the ``hard battles'', but compared to continuation 1, it is not that diverse (That’s why comparing different continuations can help to make a decision). Therefore, it gets 4 points in Diversity. In the meantime, high diversity of it also leads to some strange words in the text, and affects the overall quality. So, Continuation 4 can only get a mediocre score in Quality.\\ 
			\bottomrule
		\end{tabular}
	}
	\end{center}
		\caption{Human evaluation questionnaire for language modeling.} 
	\label{Question-LM} 
\end{table*}

\begin{table*}[t]
	\begin{center}
	\resizebox{\textwidth}{!}{
		\begin{tabular}{p{\textwidth}}
		\toprule
The goal of this review is to evaluate the quality and diversity of text paraphrase dataset. In this review, you will be given an original sentence, and its corresponding paraphrases. You should rate the paraphrases between 1 - 5 in four ways:\\
(1)	Lexical Diversity: how lexically diverse are the generated sentences?\\
(2)	Syntactical Diversity: how syntactically diverse are the generated sentences? \\
(3)	Fluency: how fluent are the generated paraphrases?\\
(4)	Relevance: how semantically consistent are between generated paraphrases and the input sentences?\\

You should score between 1 - 5, where 5 is best and 1 is worst. You can consider and make a final decision by comparing different paraphrases of the same original sentence. These sentences have been preprocessed by converting all letters to lowercase, separating punctuation, and splitting conjunctions. \textbf{Please ignore this when evaluating and do not allow it to affect your judgments.}\\ \hline
For example, if you are given the following original sentence and paraphrases:

\textbf{Original sentence}:
by adopting rules that regulate the information about the foods and their nutritional value appearing on the label , the consumers will be able to make informed and meaningful choices .

\textbf{Paraphrase 1}: \qquad Score: Lexical Diversity 5, Syntactical Diversity 5, Fluency 3, Relevance 5

the rules will be able to \textcolor{red}{adapt} food and their nutritional values listed on the labelling \textcolor{red}{of} consumers will \textcolor{blue}{be able to} be able to make informed and they are appropriate assessment .

\textbf{Paraphrase 2}: \qquad Score: Lexical Diversity 1, Syntactical Diversity 2, Fluency 1, Relevance 2

by adopting rules governing the information about food and relevance of foods and \textcolor{blue}{nutritional value} of \textcolor{blue}{nutrition value} that regulate the labelling , so that consumers .

\textbf{Paraphrase 3}: \qquad Score: Lexical Diversity 4, Syntactical Diversity 5, Fluency 1, Relevance 2

consumers can adopt rules to provide informed and nutrition value of the food and their nutritional values listed on the labelling , \textcolor{red}{consumers will be able to enable consumers} .

\textbf{Paraphrase 4}: \qquad Score: Lexical Diversity 1, Syntactical Diversity 1, Fluency 1, Relevance 1

by adopting rules that regulates the rule of food and \textcolor{blue}{their nutritional value} of food and \textcolor{blue}{their nutritional value} of \textcolor{blue}{their nutritional value} to the consumer \textcolor{red}{protection} , consumers .\\ \hline

\textbf{Analysis}: Although there are also some strange words in Paraphrase 1, we can still capture the main meaning of it. Therefore, Paraphrase 1 can get a mediocre score in Fluency and a high score in Relevance. On the other hand, Paraphrase 1 has many lexical edits and turns the original sentence into two parallel sentences, so it can full marks in both terms of Lexical and Syntactical Diversity.

Paraphrase 2 is not really finished and repeats some words in the text (highlighted in \textcolor{blue}{blue}), so it gets the lowest scores in Relevance and Fluency. Meanwhile, except for some incorrect word order transpositions, Paraphrase 2 is very similar to the original sentence. Therefore, it receives low scores in Lexical and Syntactical Diversity.

Obviously, Paraphrase 3 changes a lot lexically and syntactically. However, it is incoherent, difficult to understand (highlighted in \textcolor{red}{red}), so Paraphrase 3 scores high for Lexical and Syntactical Diversity and low for Fluency and Relevance.

Paraphrase 4 is a nonsensical text, which is not really finished and keeps repeating itself. Therefore, it gets the lowest scores from all aspects.\\
		\bottomrule
				\end{tabular}
	}
	\end{center}
		\caption{Human evaluation questionnaire for paraphrase generation.}
    \label{Question-PARA}
\end{table*}

\section{Impact Statement}
\label{impact}
Our work has developed generic generation methods to promote text diversity while maintaining comparable quality. Therefore, despite the contributes to better text generation, our proposed methods may be used to generate more human-like fake text. But the impacts are more apparent when considering deployed applications, while our proposed methods as the methodologies can not have any direct negative societal impacts. Moreover, all the datasets we used in our work are open source datasets. Wikitext-$103$ was extracted from Wikipedia, and ParaNMT-50M was created by the back-translation. Therefore, the data we used would not contain personally identifiable information or offensive content.

\end{document}